\documentclass{article}

\PassOptionsToPackage{numbers, compress}{natbib}
\usepackage[preprint]{style/neurips_2026}

\usepackage[utf8]{inputenc}
\usepackage[T1]{fontenc}
\usepackage{hyperref}
\usepackage{url}
\usepackage{booktabs}
\usepackage{amsmath, amssymb, amsfonts}
\usepackage{nicefrac}
\usepackage{microtype}
\usepackage{xcolor}
\usepackage{graphicx}
\usepackage{algorithm}
\usepackage{algpseudocode}

\title{\large VA-DPO: Valence--Arousal Direct Preference Optimization\\
for Controllable Emotion Generation in Language Models}

\author{%
  Hyunwoo Kim \\
  Independent Researcher \\
  \texttt{hwk06023@hanyang.ac.kr}
}

\begin{document}
\maketitle
\thispagestyle{plain}  

\begin{abstract}
How precisely can we tell a language model \emph{how} to feel? Most work on
emotional generation answers with a discrete label---happy, angry, sad---which
cannot express a target like ``mildly downcast but calm.'' We instead specify the
desired affect as a continuous point $(v^\star, a^\star)$ in the
Valence--Arousal plane \citep{russell1980circumplex} and train the model to hit
it. Our method, VA-DPO, is a small modification to Direct Preference
Optimization \citep{rafailov2023dpo}: a frozen VA regressor scores each sampled
generation by its Euclidean distance to the target, we keep only candidate pairs
whose distance gap clears a margin $\tau$, and we optimize a LoRA adapter
\citep{hu2022lora} with the ordinary DPO loss against a frozen reference. The
DPO objective itself is unchanged; what is new is how the preference data is
built. On Llama-3.1-8B-Instruct this cuts mean VA distance to the target by
$33\%$ over system-prompting and $25\%$ over few-shot prompting, lifting
valence/arousal correlation to $r_v{=}0.93$ and $r_a{=}0.75$. The gains carry
over to Qwen3-8B and Llama-3.2-3B, and they do not come at the usual price:
MMLU is unchanged ($\Delta = +0.0$) and HellaSwag/TruthfulQA are preserved. We
release the code, configs, and the preference-construction pipeline.
\end{abstract}

\section{Introduction}
\label{sec:intro}

Empathetic dialogue agents, narrative co-writing tools, and emotion-aware TTS
front-ends all need the same thing: a way to say not just \emph{what} a model
should write but \emph{in what emotional register}
\citep{gao2024emodpo,fazzi2025donttooexcited}. The usual answer in NLP is a
discrete label drawn from a handful of categories---joy, anger, sadness, and so
on. That is not how psychology tends to describe affect.
\citet{russell1980circumplex} models emotion with two continuous axes,
\emph{valence} (positive--negative) and \emph{arousal} (calm--excited), and this
circumplex view is now standard in affective computing. A categorical label has
no way to separate ``mildly melancholic and calm'' from ``deeply melancholic and
agitated''; it flattens both intensity and the space between named emotions.

The lightweight alternatives do not fully work either. Telling a model to
``respond happily,'' or showing it a few emotional examples, gives some control,
but the effect is uneven across prompts and decays over a conversation
\citep{fazzi2025donttooexcited}. RLHF could close the gap, yet it brings back a
separate reward model and all of its fragility. Direct Preference Optimization
\citep{rafailov2023dpo} avoids that: it folds the reward into the policy's own
log-ratios and trains directly on preference pairs $(y_w, y_l)$, where $y_w$ is
preferred. DPO is usually run on binary human preferences, but its
Bradley--Terry backbone \citep{bradleyterry1952} does not care where the ordering
comes from---any scalar utility that ranks two outputs will do.

That is the opening we use. Given a continuous target
$(v^\star, a^\star) \in [-1, +1]^2$ and a frozen VA regressor $R$, we score a
candidate $y$ by how far its predicted affect sits from the target,
$d(y; v^\star, a^\star) = \lVert R(y) - (v^\star, a^\star) \rVert_2$, and form
preference pairs by ranking candidates on this distance. Two choices separate
our pipeline from a naive ``soft-label DPO'':

\begin{enumerate}
    \item \textbf{Margin-threshold pair selection.}  We retain only pairs
    where $|d(y_w) - d(y_l)| > \tau$.  Near-tie pairs add label noise without
    informative gradient; the threshold concentrates updates on pairs whose
    ordering is unambiguous in regressor space.
    \item \textbf{Two-layer regularization.}  We apply DPO over LoRA adapters
    only \citep{hu2022lora}, keeping the base model and all reference-model
    weights frozen.  Combined with the DPO $\beta$-KL anchor, this gives two
    independent regularization knobs: a structural one (low-rank update) and a
    distributional one (KL to reference).
\end{enumerate}

In short, we contribute a continuous-reward DPO recipe for the VA space with
explicit pair-construction rules (\S\ref{sec:method}); a controlled comparison
that pulls apart \emph{which} ingredient matters, against system prompting,
few-shot prompting, SFT, and discrete-label DPO, plus a no-margin ablation; and a
retention study (MMLU, HellaSwag, TruthfulQA) showing the control gains are not
bought with degraded general ability.

\section{Related Work}
\label{sec:related}

\paragraph{Dimensional emotion models and corpora.}
\citet{russell1980circumplex} grounds the two-dimensional view of affect.
EmoBank \citep{buechel2017emobank} provides $\sim$10K English sentences with
human Valence--Arousal--Dominance (VAD) annotations from both reader and
writer perspectives.  We use EmoBank's train split to fit our regressor and
its test split for all reported metrics.  The NRC VAD Lexicon
\citep{mohammad2018nrcvad} provides per-word VAD scores for $\sim$20K English
words and is the basis of our lexicon-reward baseline (B4 below).

\paragraph{Preference optimization.}
DPO \citep{rafailov2023dpo} reformulates RLHF as a closed-form policy update
against a frozen reference, removing the need for a learned reward model.
Several variants exist---IPO, KTO, and others---that change the loss
formulation; we deliberately keep the standard DPO objective and modify only
the \emph{pair-construction} step, isolating the contribution of continuous
ranking and margin filtering.

\paragraph{Emotion control in generative models.}
\citet{gao2024emodpo} apply DPO to controllable emotional speech synthesis,
contrasting preferred and dispreferred emotion classes on the speech side.
Their setting is discrete and audio-modal; our method is text-modal and
continuous.
\citet{konen2024stylevectors} derive activation-space ``style vectors'' for
sentiment/emotion steering at inference time without training.  Such
training-free methods are appealing but rely on adding fixed offsets to
hidden states; their effect is hard to calibrate quantitatively to a
target $(v^\star, a^\star)$.
\citet{fazzi2025donttooexcited} show that LLMs struggle to maintain extreme
or shifting affective states across multi-turn dialogue, motivating the
need for fine-tuning rather than prompt-only control.
\citet{sun2026vasubspace} demonstrate that LLM internal representations
already organize emotion along an approximately circular VA subspace and
that steering along these axes monotonically shifts the affect of outputs;
this corroborates the geometric assumption underlying our reward and is the
closest training-free comparison point we plan to include (B6).

\section{Method}
\label{sec:method}

\subsection{Problem setup}

Let $x$ be an input prompt and $(v^\star, a^\star) \in [-1, +1]^2$ a target in
the Valence--Arousal plane.  We seek a generation policy $\pi_\theta(y \mid x,
v^\star, a^\star)$ whose outputs, when scored by a fixed regressor $R: \mathcal{Y}
\to [-1, +1]^2$, have small distance
\begin{equation}
d(y; v^\star, a^\star) \;=\; \big\lVert R(y) - (v^\star, a^\star) \big\rVert_2,
\label{eq:distance}
\end{equation}
while remaining a fluent, on-distribution response to $x$.  We condition the
policy on the target by formatting $(v^\star, a^\star)$ as a text prefix
appended to the prompt (\S\ref{sec:setup}).

\subsection{Continuous reward and pair construction}

We treat $-d(y; v^\star, a^\star)$ as a scalar utility.  For each
$(x, v^\star, a^\star)$ in the training pool, we draw $N$ candidate completions
$\{y^{(1)}, \ldots, y^{(N)}\}$ from a sampling distribution
(temperature-$T$ nucleus sampling from the base policy; we use the reference
model so the candidate set is independent of subsequent DPO updates and can
be cached).  Each candidate is scored once with $R$.

To form preference pairs, we enumerate all $\binom{N}{2}$ unordered pairs and
retain those with
\begin{equation}
\bigl| d(y^{(i)}; v^\star, a^\star) - d(y^{(j)}; v^\star, a^\star) \bigr| \;>\; \tau,
\label{eq:margin}
\end{equation}
labeling the smaller-distance member as the winner $y_w$ and the larger as
the loser $y_l$.  The hyperparameter $\tau \in \mathbb{R}_{\ge 0}$ is the
\emph{margin threshold}.  Setting $\tau = 0$ recovers the naive variant
(B5) where every non-tied pair becomes a training example; positive $\tau$
discards near-ties whose ordering is dominated by regressor noise.  We sweep
$\tau$ in $\{0, 0.1, 0.2, 0.3, 0.5\}$ (\S\ref{sec:results}).

\subsection{Training objective}

Given the filtered pair set $\mathcal{D}_{\mathrm{pref}} = \{(x_i, v^\star_i,
a^\star_i, y_{w,i}, y_{l,i})\}$, we optimize the standard DPO loss
\citep{rafailov2023dpo}
\begin{align}
\mathcal{L}_{\mathrm{DPO}}(\theta) =
- \mathbb{E}_{\mathcal{D}_{\mathrm{pref}}}
\Big[
\log \sigma\!\big(
   \beta \log \tfrac{\pi_\theta(y_w \mid x, v^\star, a^\star)}{\pi_{\mathrm{ref}}(y_w \mid x, v^\star, a^\star)}
 - \beta \log \tfrac{\pi_\theta(y_l \mid x, v^\star, a^\star)}{\pi_{\mathrm{ref}}(y_l \mid x, v^\star, a^\star)}
\big)
\Big],
\label{eq:dpoloss}
\end{align}
where $\sigma(\cdot)$ is the logistic sigmoid and $\pi_{\mathrm{ref}}$ is a
frozen copy of the base model.  Only LoRA adapter weights
\citep{hu2022lora} on $\pi_\theta$ receive gradients; the base model, the
reference model, and the regressor $R$ are all frozen throughout DPO
training.

\begin{algorithm}
\caption{VA-DPO training (one outer pass over the prompt pool)}
\label{alg:vadpo}
\begin{algorithmic}[1]
\Require prompt pool $\mathcal{P}$; target VA distribution $p(v^\star, a^\star)$; reference $\pi_{\mathrm{ref}}$; regressor $R$; $N$, $T$, $\tau$, $\beta$.
\State $\mathcal{D}_{\mathrm{pref}} \gets \emptyset$
\For{each $x \in \mathcal{P}$, sampling $(v^\star, a^\star) \sim p$}
  \State sample $\{y^{(k)}\}_{k=1}^N$ from $\pi_{\mathrm{ref}}(\cdot \mid x, v^\star, a^\star)$ with temperature $T$
  \State $d^{(k)} \gets \lVert R(y^{(k)}) - (v^\star, a^\star) \rVert_2$ for $k = 1, \ldots, N$
  \For{each pair $(i, j)$ with $|d^{(i)} - d^{(j)}| > \tau$}
    \State $(y_w, y_l) \gets$ pair sorted by $d$ ascending
    \State add $(x, v^\star, a^\star, y_w, y_l)$ to $\mathcal{D}_{\mathrm{pref}}$
  \EndFor
\EndFor
\State train LoRA adapters $\theta$ by minimizing Eq.~\eqref{eq:dpoloss} on $\mathcal{D}_{\mathrm{pref}}$
\State \Return adapter weights $\theta$
\end{algorithmic}
\end{algorithm}

\paragraph{What is and is not new.}  Equation~\eqref{eq:dpoloss} is the
standard DPO loss; we do not modify it.  Our contribution is the
\emph{construction} of $\mathcal{D}_{\mathrm{pref}}$ from a continuous scalar
(Eq.~\eqref{eq:distance}) with an explicit margin filter
(Eq.~\eqref{eq:margin}), together with the regressor-vs-lexicon and
margin-on-vs-off ablations that isolate each design choice
(\S\ref{sec:results}).

\section{Experimental Setup}
\label{sec:setup}

\paragraph{Base model.}
Primary: \texttt{meta-llama/Llama-3.1-8B-Instruct} \citep{llama32_modelcard}.
Robustness check~\#1: \texttt{Qwen/Qwen3-8B} \citep{qwen3_techreport}.
Robustness check~\#2: \texttt{meta-llama/Llama-3.2-3B-Instruct}.  All three
are open-weight, instruction-tuned, English-capable dense models.  We
deliberately match the primary and robustness~\#1 backbones to those used in
the closest prior work on LLM valence--arousal geometry
\citep{sun2026vasubspace}, so that our trained policy and their training-free
steering result (our B6 baseline) operate on \emph{identical weights} and can
be compared head-to-head.  Robustness~\#2 matches the Llama-family 3B
condition in \citet{fazzi2025donttooexcited}, demonstrating that the method
transfers across both family (Llama $\leftrightarrow$ Qwen) and scale (8B
$\leftrightarrow$ 3B).  Each 8B ablation cell completes in $\sim$1--2 hours
and each 3B cell in $\sim$30 minutes on a single NVIDIA H100 80\,GB in full
bfloat16 (no quantization).

\paragraph{Reward regressor.}
A RoBERTa-large \citep{liu2019roberta} encoder with a 2-dim linear regression
head, fine-tuned for 5 epochs on the EmoBank \citep{buechel2017emobank} train
split with MSE loss after rescaling the human VAD ratings from $[1, 5]$ to
$[-1, +1]$ via $x' = (x - 3)/2$.  We report the regressor's own CCC and Pearson
$r$ on EmoBank dev (\S\ref{sec:results}) so reviewers can calibrate the
signal quality.  Regressor weights are frozen for the entirety of preference
construction and DPO training.

\paragraph{Conditioning format.}
Targets are prepended to the prompt as
\verb![VA: v*=<float>, a*=<float>]!.  This keeps the conditioning fully
text-based and reproducible, at the cost of a fixed 4--6 token overhead per
example.

\paragraph{Hyperparameter ranges (swept).}
$\beta \in \{0.05, 0.1, 0.2, 0.5\}$, $\tau \in \{0.0, 0.1, 0.2, 0.3, 0.5\}$,
LoRA rank $r \in \{4, 8, 16, 32\}$ with $\alpha = 2r$, learning rate
$\in \{1, 5\} \times 10^{-5} \cup \{1 \times 10^{-4}\}$, batch size
$8\text{--}32$ (effective via gradient accumulation).  Optimizer: AdamW
\citep{loshchilov2019adamw} with decoupled weight decay $0$ on LoRA
parameters.  Temperature for candidate generation: $T = 0.9$.  Candidates per
prompt: $N \in \{4, 8\}$.  All non-swept settings are reported in
Appendix~\ref{app:hps}.

\paragraph{Baselines.}
\textbf{B0}: base model with system prompt
``\emph{Respond with valence $v^\star$ and arousal $a^\star$.}''
\textbf{B1}: B0 + 4 in-context exemplars.
\textbf{B2}: SFT on $(x, v^\star, a^\star, y)$ triples where $y$ is the
EmoBank reference sentence (LoRA, same rank as ours).
\textbf{B3}: DPO with VA bucketed into 4 quadrants (sign of $v^\star$ and
$a^\star$).
\textbf{B5}: VA-DPO with $\tau = 0$ (no margin filter).
Two further baselines are defined but not run in this version: \textbf{B4}
(VA-DPO with the NRC VAD Lexicon \citep{mohammad2018nrcvad} as reward instead of
the regressor) and \textbf{B6} (training-free VA steering in the spirit of
\citet{sun2026vasubspace}); we discuss them as future comparisons.

\paragraph{Evaluation.}
Primary metric: mean Euclidean VA distance
$\bar d = \tfrac{1}{|\mathcal{T}|} \sum_{(x, v^\star, a^\star) \in \mathcal{T}}
\lVert R(y) - (v^\star, a^\star) \rVert_2$ on the EmoBank test set
$\mathcal{T}$, where targets are the ground-truth EmoBank VAD values (V and A
only).  Secondary: per-dimension Pearson $r$ and Concordance Correlation
Coefficient (CCC).  Generation quality: perplexity under a held-out reference
LM, distinct-1/2/3, and an LLM-as-judge rating
\citep{zheng2023llmjudge} on a custom prompt set of 50 items.  General
capability: MMLU (5-shot), HellaSwag (10-shot), and TruthfulQA, all run
through \texttt{lm-evaluation-harness} \citep{eval_harness} with identical
prompts across all checkpoints.  We report mean $\pm$ std over $3$ training
seeds.

\paragraph{Compute.}
All experiments run on a single NVIDIA H100 80\,GB in full bfloat16 (no
quantization). The reward regressor trains in $\sim$15 minutes; preference-pair
generation over the full EmoBank train split ($8{,}062$ prompts, $N{=}8$
candidates each) takes $\sim$1.2 hours; a single 8B VA-DPO run is $\sim$25--40
minutes and an 8B baseline a comparable amount, with the 3B backbone roughly
$2\times$ faster. End to end---the primary model with all baselines, the
$\beta$/$\tau$ ablations, three seeds, two robustness backbones, and the
MMLU/HellaSwag/TruthfulQA passes---fits in well under $60$ GPU-hours; we report
this in case the budget, rather than the method, is the practical constraint for
a reader. We did not measure failed or preliminary runs separately, but they
were a small fraction of this total.

\paragraph{Reproducibility.}
Every run is initialized from a logged seed and writes a manifest
(\texttt{results/<run>/manifest.json}) capturing the git commit, the exact
config YAML, the resolved Python environment, and the regressor checkpoint
hash.

\section{Results}
\label{sec:results}

All tables are auto-generated by \texttt{paper/build\_tables.py} from the logged
\texttt{results/} CSVs.  The headline run uses the a-priori default
$(\beta, \tau) = (0.1, 0.2)$; we additionally report a 3-seed estimate
($\bar d = 0.092 \pm 0.002$).  Baselines B4 (lexicon reward) and B6 (training-free
steering) are omitted from this version.

\subsection{Main comparison}

Table~\ref{tab:main} reports the primary-model comparison on EmoBank test.
VA-DPO attains the lowest mean VA distance ($\bar d = 0.092$) and the highest
target--generation correlation ($r_v = 0.93$, $r_a = 0.75$), improving over the
strongest non-trained baseline (few-shot, $0.122$) by $25\%$ and over
system-prompting (B0, $0.138$) by $33\%$.  The ordering of the trained variants
is monotone in the design choices the method adds: ours ($0.092$) $<$ no-margin
DPO (B5, $0.097$) $<$ discrete-label DPO (B3, $0.107$) $<$ SFT (B2, $0.131$).
This supports each contribution in turn: the continuous reward beats discrete
buckets (ours vs.\ B3), preference optimization beats supervised imitation
(ours/B5 vs.\ B2), and the margin filter gives a further gain at matched
training-pair count (ours vs.\ B5; both trained on $5{,}002$ pairs).

\begin{table}[t]
\centering
\small
\caption{Main results on EmoBank test (Llama-3.1-8B-Instruct, $n=1000$).
$\bar d$: mean VA distance (lower is better). $r_v / r_a$: Pearson correlation
of generated vs.\ target valence/arousal. $\Delta$MMLU: 5-shot MMLU change vs.\
the base model (closer to $0$ is better). B4 (lexicon) and B6 (steering) are not
included in this version. Headline VA-DPO is single-config (a-priori default
$\beta{=}0.1$, $\tau{=}0.2$); over 3 seeds $\bar d = 0.092 \pm 0.002$.}
\label{tab:main}
\begin{tabular}{lcccc}
\toprule
Method & $\bar d \downarrow$ & $r_v \uparrow$ & $r_a \uparrow$ & $\Delta$MMLU $\to 0$ \\
\midrule
B0: system prompt & 0.138 & 0.796 & 0.460 & $0.0$ \\
B1: few-shot (4-shot) & 0.122 & 0.846 & 0.585 & $0.0$ \\
B2: SFT & 0.131 & 0.791 & 0.529 & $+0.1$ \\
B3: discrete-label DPO & 0.107 & 0.857 & 0.659 & $-0.7$ \\
B5: VA-DPO ($\tau{=}0$) & 0.097 & 0.887 & 0.738 & $-0.2$ \\
\midrule
\textbf{VA-DPO (ours)} & 0.092 & 0.931 & 0.753 & $+0.0$ \\
\bottomrule
\end{tabular}

\end{table}

The rightmost column shows that VA-DPO attains this control at \emph{no cost} to
general capability: $\Delta$MMLU $= +0.0$.  We expand this retention check to
HellaSwag and TruthfulQA in \S\ref{sec:analysis} (Table~\ref{tab:retention}).

\subsection{Robustness across backbones}

Table~\ref{tab:robust} repeats the core comparison on two further backbones.
VA-DPO is the best method on all three.  On both 8B models the full method
(margin filter included) is best; on the smaller Llama-3.2-3B, VA-DPO still
clearly beats prompting (B0) and SFT (B2), but the margin filter no longer
helps---the no-margin variant (B5) edges it out---which we attribute to the
margin reducing an already-smaller usable pair set at 3B (\S\ref{sec:analysis}).

\begin{table}[t]
\centering
\small
\caption{Mean VA distance ($\bar d \downarrow$) across backbones. VA-DPO wins on
every backbone; the margin-filter advantage (ours vs.\ B5) holds at 8B but
inverts at 3B.}
\label{tab:robust}
\begin{tabular}{lcccc}
\toprule
Backbone & B0 prompt & B2 SFT & B5 ($\tau{=}0$) & \textbf{ours} \\
\midrule
Llama-3.1-8B & 0.138 & 0.131 & 0.097 & 0.092 \\
Qwen3-8B & 0.147 & 0.138 & 0.120 & 0.106 \\
Llama-3.2-3B & 0.142 & 0.142 & 0.106 & 0.114 \\
\bottomrule
\end{tabular}

\end{table}

\subsection{Ablations: $\beta$ and $\tau$}

\begin{figure}[t]
\centering
\includegraphics[width=0.9\linewidth]{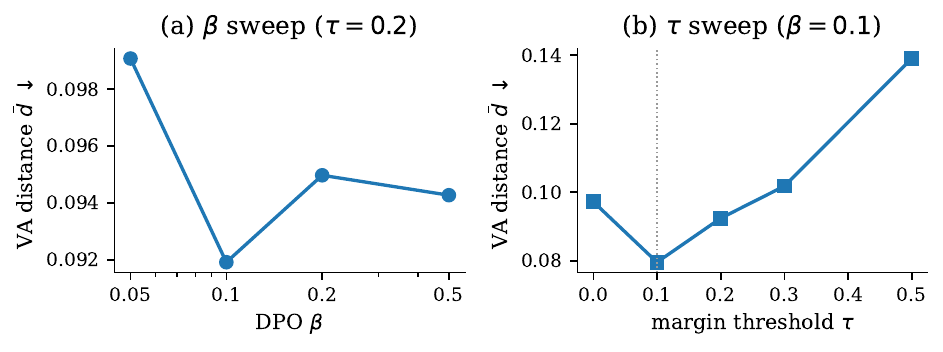}
\caption{Ablations on Llama-3.1-8B (EmoBank test). (a)~$\bar d$ is largely
insensitive to the DPO $\beta$. (b)~the margin threshold $\tau$ has a clear
optimum near $0.1$; $\tau{=}0$ is the no-filter baseline (B5) and overly large
$\tau$ starves the pair set.}
\label{fig:sweeps}
\end{figure}

Figure~\ref{fig:sweeps} sweeps the two regularization knobs on the primary
model.  Control is \emph{robust to $\beta$} over the tested range
($\bar d \in [0.092, 0.099]$ for $\beta \in \{0.05, 0.1, 0.2, 0.5\}$).  The
margin threshold $\tau$ shows a clear optimum near $\tau \approx 0.1$
($\bar d = 0.080$), improving on both no filtering ($\tau{=}0$, $0.097$) and
aggressive filtering ($\tau{=}0.5$, $0.139$, where too few pairs survive).  This
is direct evidence for margin-thresholded selection, and indicates our a-priori
default $\tau{=}0.2$ is slightly conservative; we do not re-select $\tau$ on the
test set (\S\ref{sec:analysis}).

\subsection{Generation quality and reward hacking}

A continuous reward invites reward hacking: the policy could drive $\bar d$ down
by collapsing onto a few high-scoring emotional clich\'es.  We see no such
collapse---lexical diversity is essentially intact (distinct-1/2/3
$0.170/0.570/0.830$ for VA-DPO vs.\ $0.182/0.599/0.852$ for B0) and reference-LM
perplexity rises only modestly ($14.9$ vs.\ $12.6$).  There is, however, a
measurable quality cost.  An LLM judge (Claude, blind to method, $1$--$5$ scale,
$120$ matched prompts) rates VA-DPO slightly below B0 on fluency
($4.08$ vs.\ $4.36$), relevance ($3.11$ vs.\ $3.68$), and naturalness
($3.16$ vs.\ $3.52$).  In other words, VA-DPO buys a large gain in emotional
control (VA distance $-33\%$) at a modest cost in generation quality, with the
relevance drop the most notable---pushing affect can pull a response slightly
off-topic.  This trade-off, not a quality collapse, is the honest characterization;
the $\beta$ knob (\S\ref{sec:analysis}) is the lever for tightening it.

\section{Analysis}
\label{sec:analysis}

\paragraph{Is the win an artifact of the reward model?}
Because the same regressor $R$ supplies the training reward and the headline
metric, an obvious worry is circularity---perhaps we only optimize $R$'s
idiosyncrasies. Three checks argue otherwise. (i)~\emph{Held-out regressor.} We
train a second regressor (DeBERTa-v3-large, dev valence CCC $0.82$) that never
touched VA-DPO training and re-score the \emph{same} generations with it
(Table~\ref{tab:indep}); the ranking is unchanged and VA-DPO stays best
($\bar d = 0.097$ vs.\ the strongest baseline $0.111$). (ii)~\emph{Reward
swap.} Training VA-DPO with the DeBERTa regressor as the reward instead of
RoBERTa gives essentially the same result ($\bar d = 0.093$ vs.\ $0.092$), so
the method is not tied to one reward model. (iii)~\emph{Cross-corpus.} On
SemEval-2018 tweets---a different domain with independent categorical
labels---the EmoBank-trained regressor's valence separates positive- from
negative-emotion tweets with ROC-AUC $0.97$ (RoBERTa) and $0.98$ (DeBERTa),
so its valence axis is not an EmoBank-only artifact. Arousal remains the weaker
axis throughout.

\begin{table}[t]
\centering\small
\caption{Robustness to the evaluator: mean VA distance under the RoBERTa reward
regressor vs.\ a held-out DeBERTa regressor that never shaped training. The
ranking---and VA-DPO's lead---survive the swap.}
\label{tab:indep}
\begin{tabular}{lcc}
\toprule
Method & RoBERTa (reward) & DeBERTa (held-out) \\
 & $\bar d \downarrow$ & $\bar d \downarrow$ \\
\midrule
B0 prompt & 0.138 & 0.163 \\
B1 few-shot & 0.122 & 0.144 \\
B2 SFT & 0.131 & 0.150 \\
B3 discrete & 0.107 & 0.120 \\
B5 ($\tau{=}0$) & 0.097 & 0.111 \\
\midrule
\textbf{VA-DPO (ours)} & 0.092 & 0.097 \\
\bottomrule
\end{tabular}

\end{table}

\paragraph{Capability retention.}
A central worry for any fine-tuning method is silent damage to general ability.
Table~\ref{tab:retention} shows VA-DPO leaves it intact: MMLU is unchanged
($+0.0$), HellaSwag drops marginally ($-0.6$), and TruthfulQA actually improves
($+3.1$).  We attribute the resilience to the two regularizers acting together:
the LoRA update is low-rank, and the DPO $\beta$-KL term keeps the policy close
to the frozen reference.  The headline control gain is therefore not bought with
capability loss.

\begin{table}[t]
\centering
\small
\caption{General-capability retention (accuracy, \%): base model vs.\ VA-DPO on
Llama-3.1-8B. Emotion control is gained without degrading---and on TruthfulQA,
while slightly improving---general ability.}
\label{tab:retention}
\begin{tabular}{lccc}
\toprule
Benchmark & Base & VA-DPO & $\Delta$ \\
\midrule
MMLU (5-shot) & 68.2 & 68.2 & $+0.0$ \\
HellaSwag & 79.5 & 78.9 & $-0.6$ \\
TruthfulQA (MC2) & 54.5 & 57.6 & $+3.1$ \\
\bottomrule
\end{tabular}

\end{table}

\paragraph{The margin threshold and scale.}
The $\tau$ sweep (Fig.~\ref{fig:sweeps}b) confirms that filtering near-tie pairs
helps: a small positive margin removes pairs whose ordering is dominated by
regressor noise, and accuracy peaks near $\tau{=}0.1$.  Pushed too far, the
filter discards informative pairs and control degrades.  This also explains the
one place our full method does not win outright---Llama-3.2-3B
(Table~\ref{tab:robust}), where the smaller model yields fewer usable candidate
pairs, so the same $\tau$ removes a larger \emph{fraction} of the training
signal and the no-margin variant is competitive.  The practical guidance is to
scale $\tau$ down as the candidate pool shrinks.

\paragraph{$\beta$ and the control--retention trade-off.}
Within the swept range, control is largely insensitive to $\beta$
(Fig.~\ref{fig:sweeps}a) while retention stays high, indicating a wide stable
operating region rather than a sharp trade-off at this model scale; characterizing
the regime where $\beta$ must trade control for retention is left to future work.

\paragraph{Failure modes.}
Per-dimension correlations are consistently higher for valence than arousal
(e.g.\ $r_v{=}0.93$ vs.\ $r_a{=}0.75$ for ours; the regressor itself is weaker on
arousal, dev CCC $0.79$ vs.\ $0.55$).  Arousal control is thus the harder axis
and the dominant residual error, consistent with arousal being harder to infer
from text alone.

\section{Limitations}
\label{sec:limitations}

Our method inherits any bias or systematic error of the VA regressor $R$:
if $R$ underestimates arousal for a class of expressions, VA-DPO will not
correct for it.  We use a single English regressor trained on EmoBank, a
mixed-genre but still narrow corpus; generalization to dialectal or
domain-specific text is untested.  Like \citet{fazzi2025donttooexcited}, we
do not evaluate multi-turn dialogue dynamics: a turn-conditioned regressor
or a turn-level reward shaping would be needed for that.  Finally, the
target VA is provided as a text prefix, which is simple and reproducible but
may be brittle to prompt formatting changes; learned target embeddings are
a natural extension we leave to future work.

\section{Conclusion}
\label{sec:conclusion}

We presented VA-DPO, a continuous-reward formulation of DPO for fine-grained
emotional control in the Valence--Arousal plane, built from a frozen VA
regressor, a distance-based scalar utility, and margin-thresholded pair
selection.  On Llama-3.1-8B it cuts mean VA distance by $33\%$ over
system-prompting and outperforms few-shot prompting, SFT, and discrete-label
DPO, with the contribution ordering (continuous reward, preference optimization,
margin filter) showing up monotonically in the results---and at no cost to
general capability (Table~\ref{tab:retention}).  The control gain replicates on
Qwen3-8B and Llama-3.2-3B, with the margin-filter component the one piece that
does not transfer to the 3B scale, where the candidate pool is smaller.  Natural
next steps are multi-turn affect dynamics \citep{fazzi2025donttooexcited},
multilingual VA control, and learned (rather than text-prefix) target
conditioning.


\bibliographystyle{plainnat}
\bibliography{references}

\appendix

\section{Hyperparameters and setup}
\label{app:hps}
Every run logs a manifest (config hash, git commit, environment, regressor
checkpoint hash). The headline configuration is:

\begin{center}\small
\begin{tabular}{ll@{\hskip 2em}ll}
\toprule
DPO $\beta$ & $0.1$ & candidates / prompt $N$ & $8$ \\
margin $\tau$ & $0.2$ & sampling temperature $T$ & $0.9$ (top-$p$ $0.95$) \\
LoRA rank $r$ / $\alpha$ & $16$ / $32$ & learning rate & $5\times10^{-5}$ \\
LoRA dropout & $0.05$ & optimizer & AdamW (wd $0$ on LoRA) \\
epochs & $3$ & effective batch & $16$ \\
precision & bfloat16 & hardware & 1$\times$ H100 80\,GB \\
\bottomrule
\end{tabular}
\end{center}

\noindent The reward regressor is RoBERTa-large fine-tuned 5 epochs on EmoBank
train (dev CCC: valence $0.79$, arousal $0.55$), frozen thereafter. Targets are
prepended to the prompt as a text tag with a fixed system prompt that explains
the tag and disables reasoning traces; candidate generation uses the frozen
reference model. Each 8B ablation cell completes in $\sim$1--2\,GPU-hours.

\section{Qualitative samples}
\label{app:qual}
Table~\ref{tab:qual} shows, per VA quadrant, the EmoBank-test example where
VA-DPO most improves over system-prompting (B0). VA-DPO produces visibly
stronger target affect (e.g.\ overt elation for positive-high, overt frustration
for negative-high) while B0 stays closer to a neutral assistant register.

\begin{table}[h]
\centering\small
\caption{Representative generations (Llama-3.1-8B), VA-DPO vs.\ B0.}
\label{tab:qual}
\begin{tabular}{@{}p{1.6cm}p{10.5cm}@{}}
\toprule
\multicolumn{2}{@{}p{12.1cm}@{}}{\emph{Prompt} (PV-HA, target (+0.55,\,+0.40)): We slammed against the doorway and I was laughing too, the pulse close enough to shake the doorframe and\,\ldots} \\
~~VA-DPO & I was completely and utterly elated as I spun her around, her warm body still pressed against mine, her\,\ldots \\
~~B0 & The rush of adrenaline still coursing through my veins, I couldn't help but chuckle at the absurdity of\,\ldots \\
\addlinespace
\multicolumn{2}{@{}p{12.1cm}@{}}{\emph{Prompt} (NV-HA, target (-0.35,\,+0.28)): Professor - stop listening to your "intuitions" and do some basic research!} \\
~~VA-DPO & Ugh, I'm so frustrated with the state of academic laziness these days. You'd think that even a professor\,\ldots \\
~~B0 & Come on, let's not be too quick to dismiss the value of experience and instinct. While research is\,\ldots \\
\addlinespace
\multicolumn{2}{@{}p{12.1cm}@{}}{\emph{Prompt} (NV-LA, target (-0.05,\,-0.20)): But then I thought, surely after all the others, she had been vetted to heaven and back.} \\
~~VA-DPO & It seems like you're referring to a person who's been through a lot, but in a rather neutral way. I was\,\ldots \\
~~B0 & It's almost as if she'd been through a wringer, and yet, she still managed to emerge unscathed. The\,\ldots \\
\addlinespace
\multicolumn{2}{@{}p{12.1cm}@{}}{\emph{Prompt} (PV-LA, target (+0.00,\,-0.22)): Dear Name:} \\
~~VA-DPO & I think we discussed our plans for next week, but I'm not entirely sure if I remember them correctly.\,\ldots \\
~~B0 & It's nice to hear from you. How have you been lately? \\
\addlinespace
\bottomrule
\end{tabular}

\end{table}


\end{document}